\documentclass[10pt,twocolumn,letterpaper]{article}

\usepackage{hhline}
\usepackage{wacv}
\usepackage{subcaption}
\usepackage{graphics} 
\usepackage{epsfig} 
\usepackage{mathptmx} 
\usepackage{times} 
\usepackage{amsmath} 
\usepackage{amssymb}  
\usepackage{booktabs}
\usepackage{multirow}
\usepackage{cleveref}
\usepackage{afterpage}
\usepackage{dblfloatfix}    

\wacvfinalcopy 

\def\wacvPaperID{111} 

\ifwacvfinal\fi
\begin{document}

\title{MultiNet: Multi-Modal Multi-Task Learning for Autonomous Driving}

\author{Sauhaarda Chowdhuri \\
	Westview High School\\
	{\tt\small sauhaarda@gmail.com}
	\and
	Tushar Pankaj \\
	Berkeley DeepDrive Center\\
	{\tt\small tpankaj@berkeley.edu}
	\and
	Karl Zipser \\
	Berkeley DeepDrive Center\\
	{\tt\small karlzipser@berkeley.edu}
}

\maketitle
\ifwacvfinal\thispagestyle{empty}\fi

\begin{abstract}
	Autonomous driving requires operation in different behavioral modes ranging from lane following and intersection 
crossing to turning and stopping. However, most existing deep learning approaches to autonomous driving do not consider the behavioral mode in the training strategy.
This paper describes a technique for learning multiple distinct behavioral modes in a single deep neural network through the use of multi-modal multi-task learning.
We study the effectiveness of this approach, denoted MultiNet, using self-driving model cars for driving in unstructured environments such as sidewalks and unpaved roads. Using labeled data from over one hundred hours of driving our fleet of 1/10th scale model cars,
we trained different neural networks to predict the steering angle and driving speed of the vehicle 
in different behavioral modes. 
We show that in each case, MultiNet networks outperform networks trained on individual modes while using a fraction of the total number of parameters.
\end{abstract}

\section{INTRODUCTION}
\label{sec:intro}

Most research on driving with DNNs has focused on the single task of steering prediction. \cite{bojarski2016end,muller2006off,chen2015deepdriving}. We consider these approaches as \textit{Single Task Learning} (STL), as they focus on training to perform an individual task.

Multi-task learning (MTL) research has shown that training on side tasks related to the main operation of a deep neural network can enhance its learning capabilities \cite{zhang2012convex, zhang2014facial, evgeniou2004regularized, argyriou2007multi}. These auxiliary tasks, such as lane or vehicle detection \cite{DBLP:journals/corr/HuvalWTKSPARMCM15}, enhance the quality of the training, resulting in improved performance on the primary task \cite{caruana1998multitask}.

Additional research is being conducted on multi-modal learning\cite{DBLP:journals/corr/abs-1801-06734, DBLP:journals/corr/KaiserGSVPJU17, eitel2015multimodal}. Multi-modal learning involves relating information from multiple types of input. For example, a single network which handles both audio and video data inputs \cite{ngiam2011multimodal} would be multi-modal. Multi-modal approaches fuse  multiple inputs into a shared representation at some stage of the network either through a deep autoencoder \cite{ngiam2011multimodal} or concatenation \cite{eitel2015multimodal}. This shared representation is then processed by a late fusion network to produce the desired output.

\begin{figure}[t]
	\centering
	\includegraphics[width=\columnwidth]{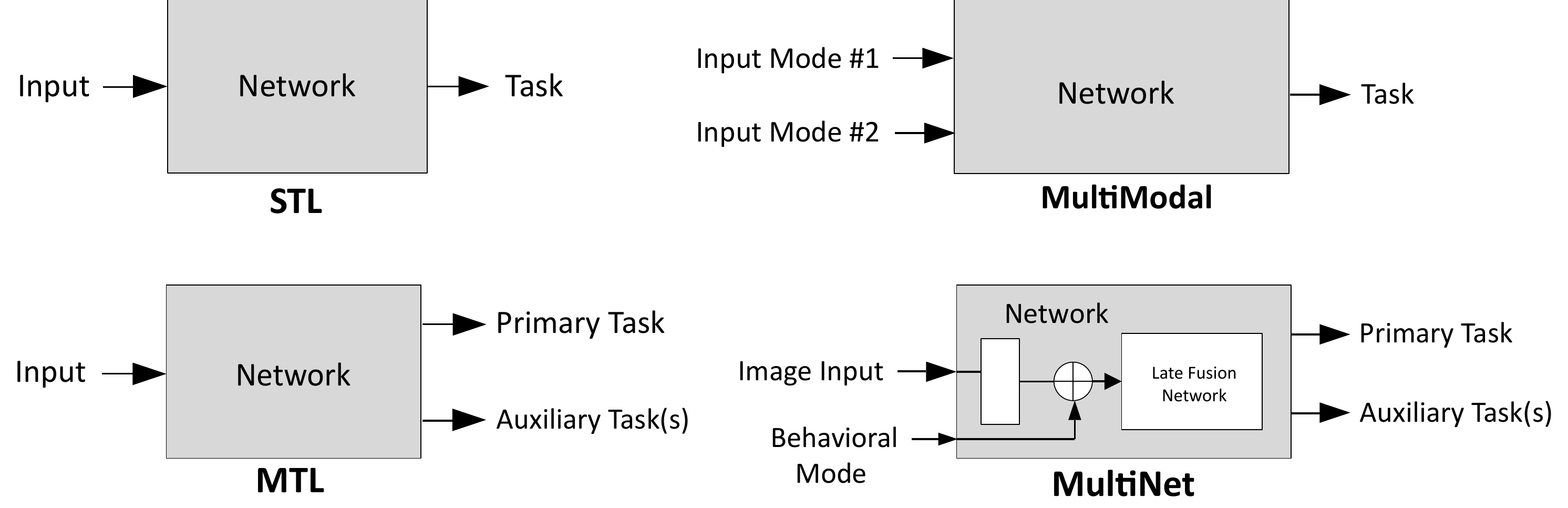}
	\caption{Training Styles}
	\label{fig:moment}
\end{figure}

\begin{table*}
	\centering
	\begin{tabular}{|l|l|l|l|l|l|}
		\hline
		\textbf{}             & \textbf{Ours} & \textbf{KITTI} & \textbf{Cityscapes} & \textbf{ApolloScape} & \textbf{Mapillary} \\ \hline
		\# Images             & 11,000,000    & 14,999         & 5000 (+2000)        & 143,906              & 25,000             \\ \hline
		Multiple Cities       & \textbf{Yes}  & \textbf{Yes}   & \textbf{Yes}                  & \textbf{Yes}                   & \textbf{Yes}       \\ \hline
		Multiple Weathers     & \textbf{Yes}  & No             & \textbf{Yes}                  & \textbf{Yes}                   & \textbf{Yes}       \\ \hline
		Multiple Times of Day & \textbf{Yes}  & No             & No                  & \textbf{Yes}                   & \textbf{Yes}       \\ \hline
		Corrective Data               & \textbf{Yes}  & No             & No                  & No                   & No                 \\ \hline
		Behavioral Modes               & \textbf{Yes}  & No             & No                  & No                   & No                 \\ \hline
	\end{tabular}
	\caption{Comparison to On-road Datasets}
	\label{tab:comp}
\end{table*}

In this paper, we propose a new method for unstructured autonomous driving in multiple behavioral modes by combining multi-modal learning with MTL. In this method, the MTL auxiliary tasks consist of additional inferred speed and steering values which form a planned trajectory. We introduce multiple distinct driving behaviors, or \textit{behavioral modes}, in which the vehicle can operate.

Our main innovation is using the behavioral mode as a second type of input to the network, which is concatenated with the input image processing stream to allow for separate driving behaviors to form within a single multi-modal network.

We also present our unique dataset of 1/10th scale model cars driving in unstructured conditions e.g.\ sidewalks, trails, and unpaved roads. A sidewalk driving dataset is relevant for solving the last mile problem in delivery, for which self-driving model cars have been cited as a viable solution \cite{bascelli2018sidewalk}. Additionally, the small size of the model cars allows for safe experiments with atypical driving behaviors and the collection of valuable data involving the vehicle making and recovering from mistakes. We make use of this live corrective data in our implementation of a novel approach to the DAgger algorithm described in \Cref{sec:dag}.

The concurrent work of \cite{intel_paper} presents a multi-task multi-modal approach for autonomous driving on model cars which makes use of a high-level directional command to direct the model car to turn left, right, or go straight. However, the directional command is simply used to select between separate MTL networks trained on specific commands rather than fused into the processing stream of the network. This approach does not scale to a greater number of high level commands as the number of sub networks is proportional to the number of input commands. Since the sub-networks used do not utilize information from the higher level command, these networks are analogous to our baseline MTL networks which are shown to be outperformed by a single MultiNet network.

Another recent work \cite{DBLP:journals/corr/abs-1801-06734} also presents a multi-task multi-modal approach to autonomous driving on the road using the secondary input of past inferred driving speeds. Thus, while this is multi-modal learning, it is not directly comparable to our approach in which higher level information is inserted into the network. Additionally, the network structure used is fundamentally different. Their model combines a single STL network for steering prediction with a multi-modal network for speed prediction. This soft parameter sharing approach has been shown to be susceptible to overfitting in comparison to hard parameter sharing \cite{DBLP:journals/corr/Ruder17a}.

This paper is organized as follows.
Section \ref{sec:dataset} describes details of our dataset and how it compares with other standard data sets.
Section \ref{sec:approach} describes the specific innovations of the MultiNet approach and introduces our own deep convolutional neural network, \textit{Z2Color}.
Section \ref{sec:experiments} covers the experiments conducted through evaluation of network validation loss for different behavioral modes, as well as evaluation in on-the-road tests.
Finally, Section \ref{sec:conclusion} summarizes the major contributions of this paper and suggests areas for future work.

\afterpage{
\begin{figure*}[!th]
\centering
\includegraphics[width=0.9\textwidth]{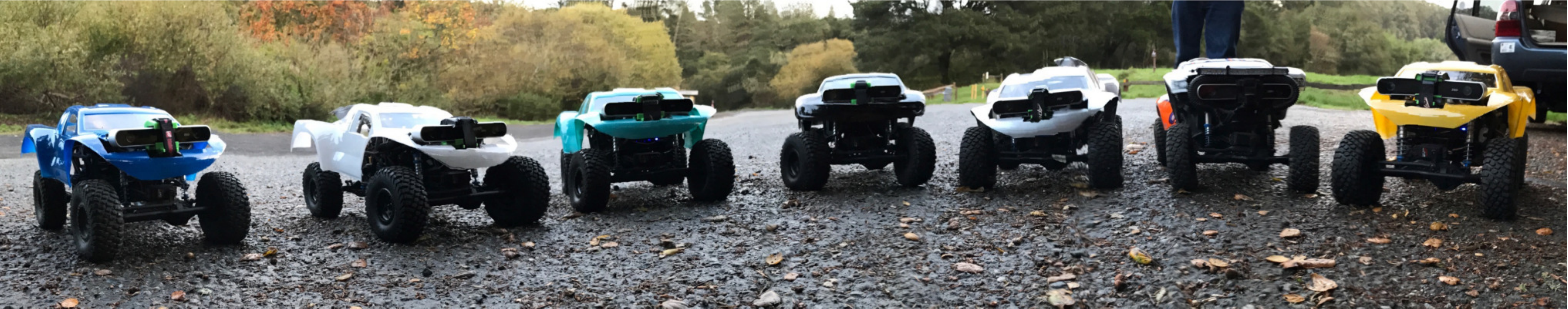}
\caption{Fleet of Model Cars}
\label{fig:fleet}
\end{figure*}
\begin{figure*}[!th]
    \centering
     \begin{subfigure}{0.3\textwidth}
       \centering
       \includegraphics[width=\linewidth]{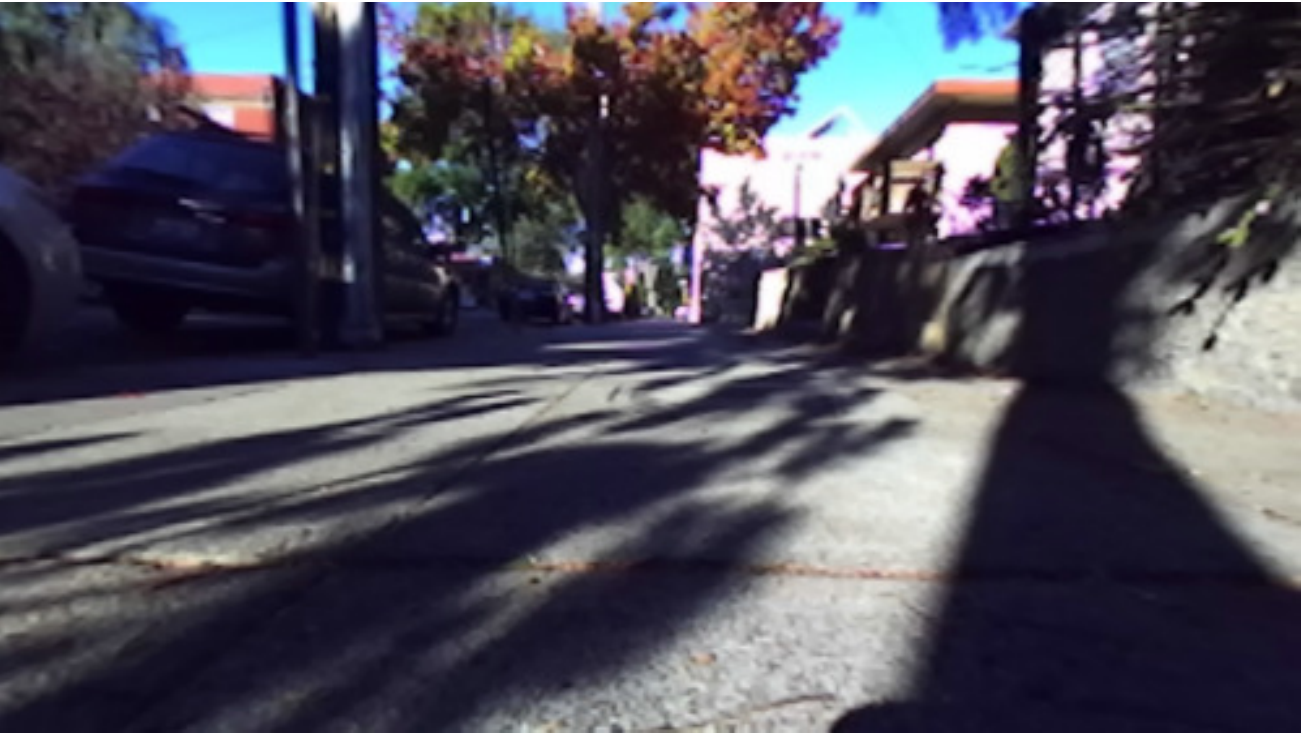}
       \caption{Day Time}
     \end{subfigure}
     \begin{subfigure}{0.3\textwidth}
       \centering
       \includegraphics[width=\linewidth]{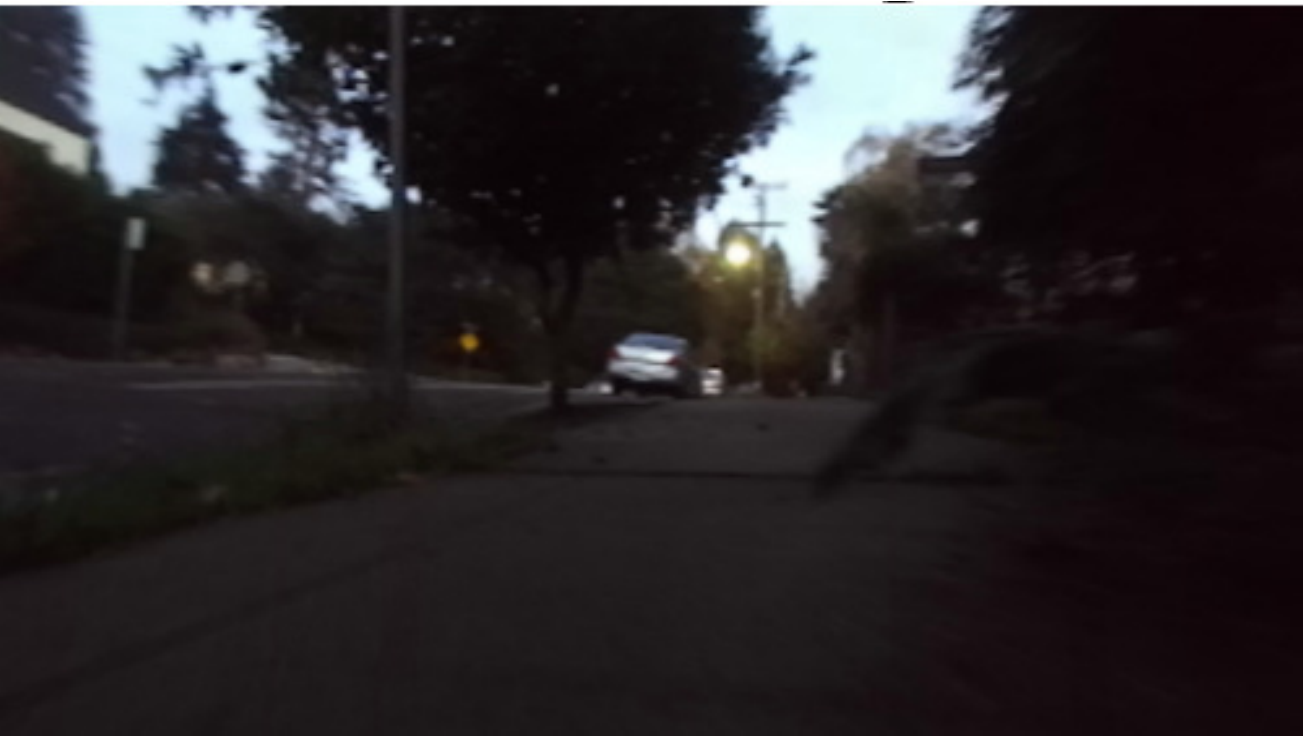}
       \caption{Evening}
     \end{subfigure}
     \begin{subfigure}{0.3\textwidth}
       \centering
       \includegraphics[width=\linewidth]{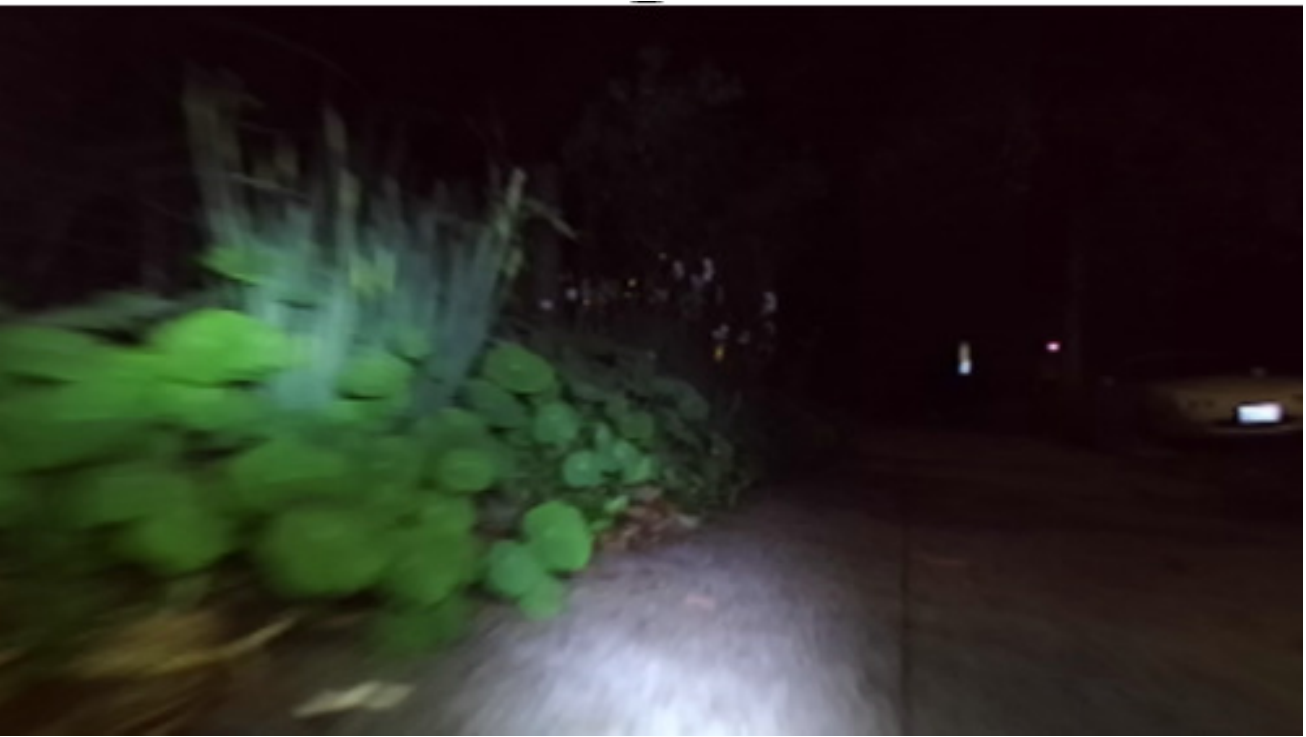}
       \caption{Night Time}
     \end{subfigure}
        \begin{subfigure}{0.3\textwidth}
       \centering
       \includegraphics[width=\linewidth]{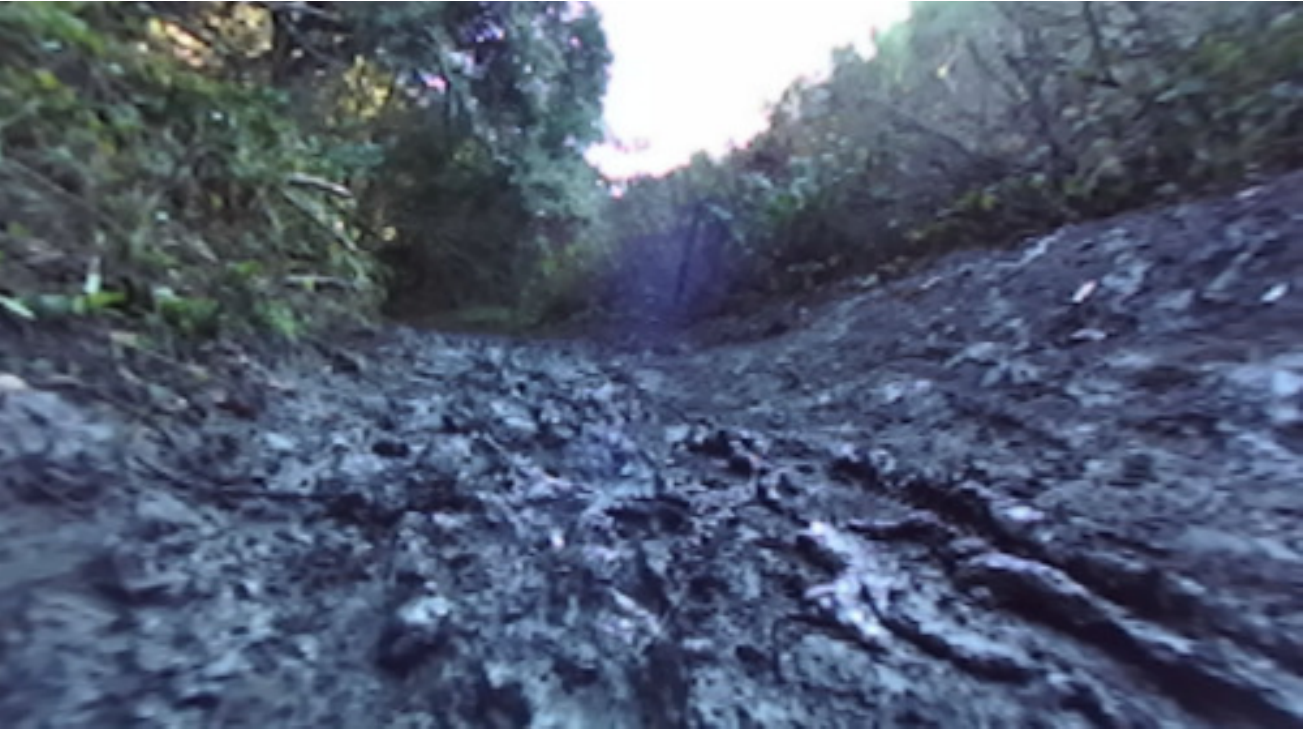}
       \caption{Muddy Area}
    \end{subfigure}
    \begin{subfigure}{0.3\textwidth}
       \centering
       \includegraphics[width=\linewidth]{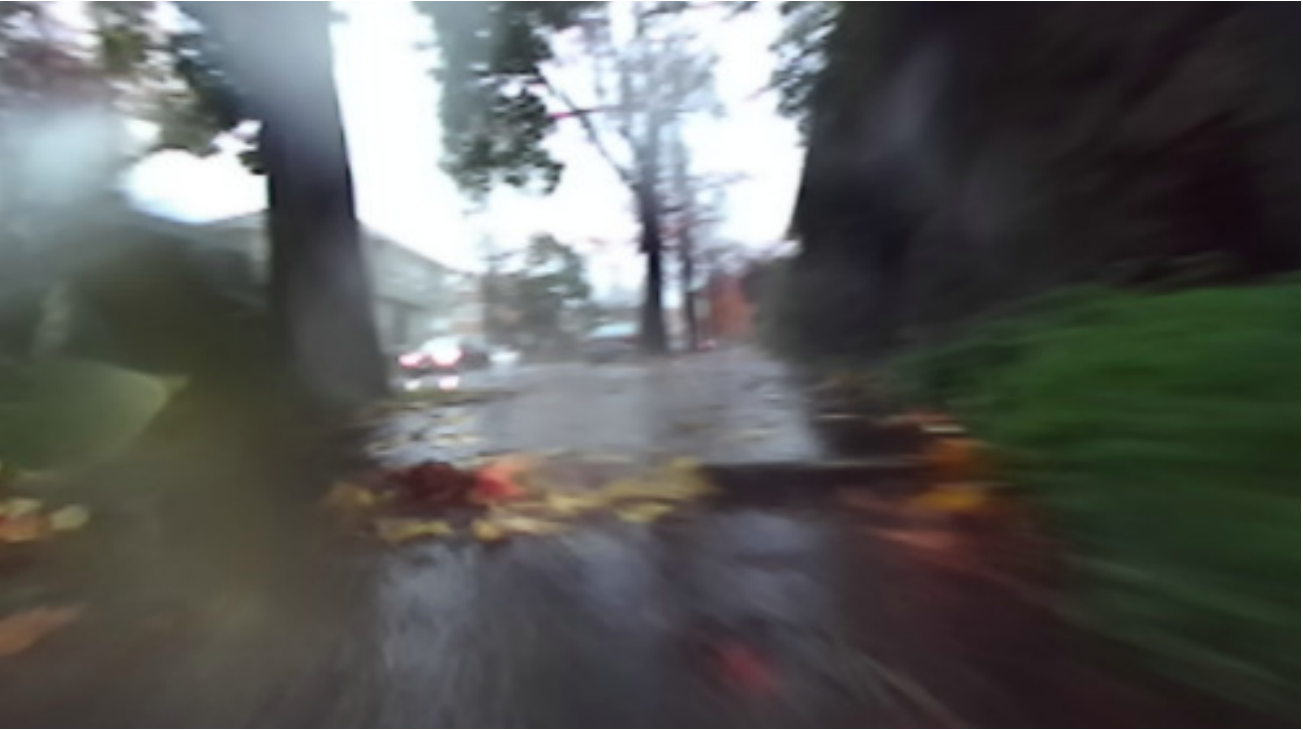}
       \caption{Rainy Area}
    \end{subfigure}
    \begin{subfigure}{0.3\textwidth}
       \centering
       \includegraphics[width=\linewidth]{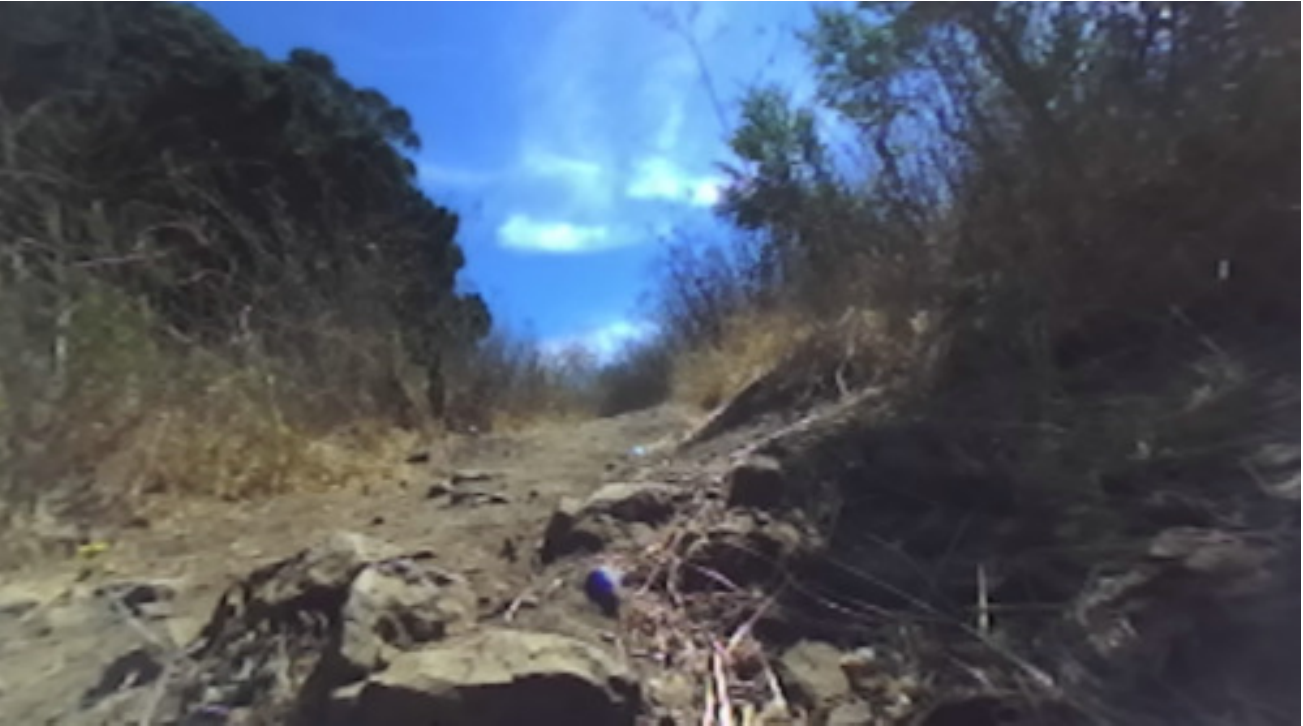}
       \caption{Bumpy Area}
    \end{subfigure}
    \caption{Diverse Conditions in Dataset}
    \label{fig:diverse}
\end{figure*}

\begin{figure*}[!th]
    \centering
    \begin{subfigure}{0.3\textwidth}
       \centering
       \includegraphics[width=\linewidth]{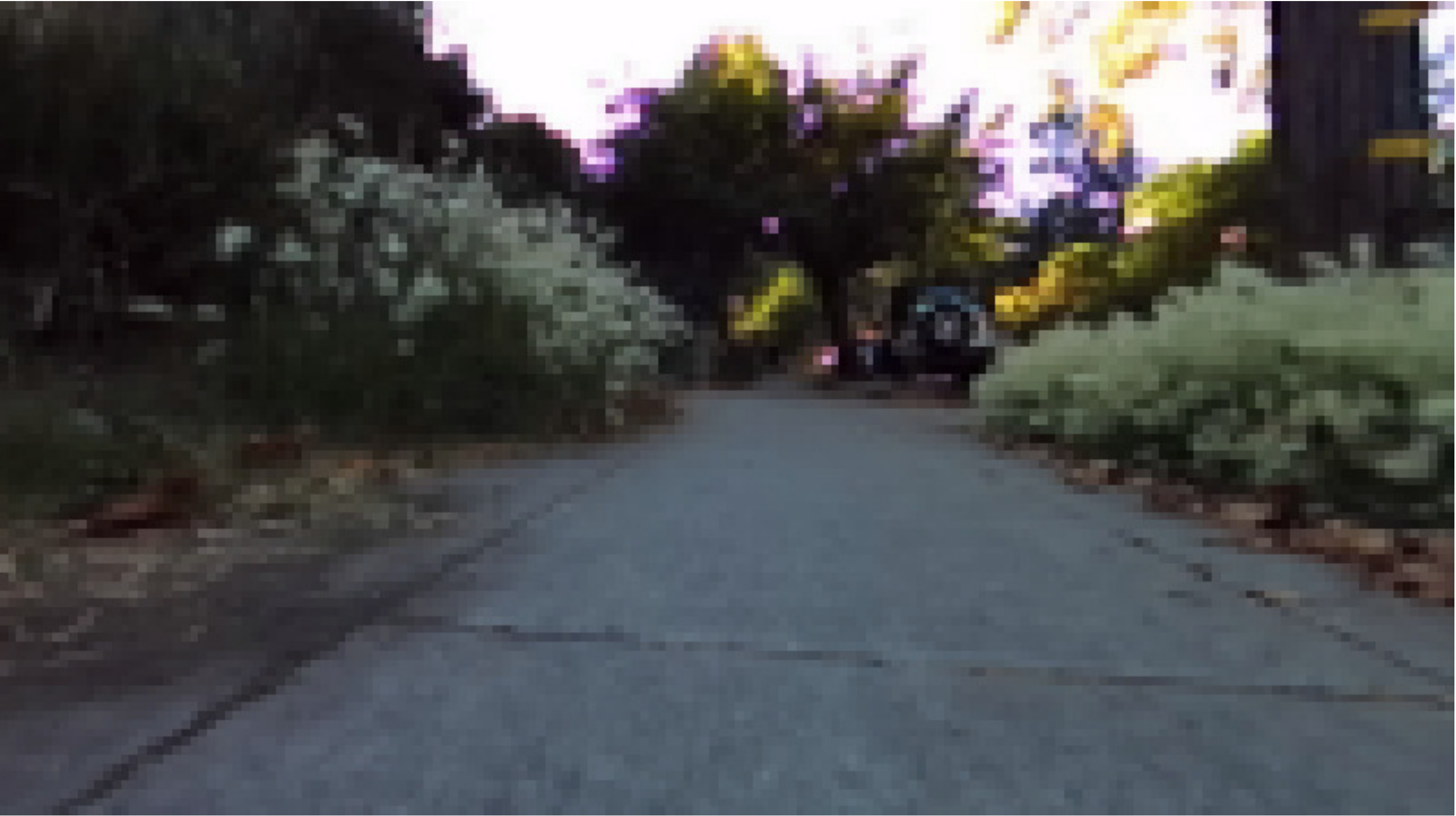}
       \caption{Direct Mode}
       \label{fig1:direct}
    \end{subfigure}
    \begin{subfigure}{0.3\textwidth}
       \centering
       \includegraphics[width=\linewidth]{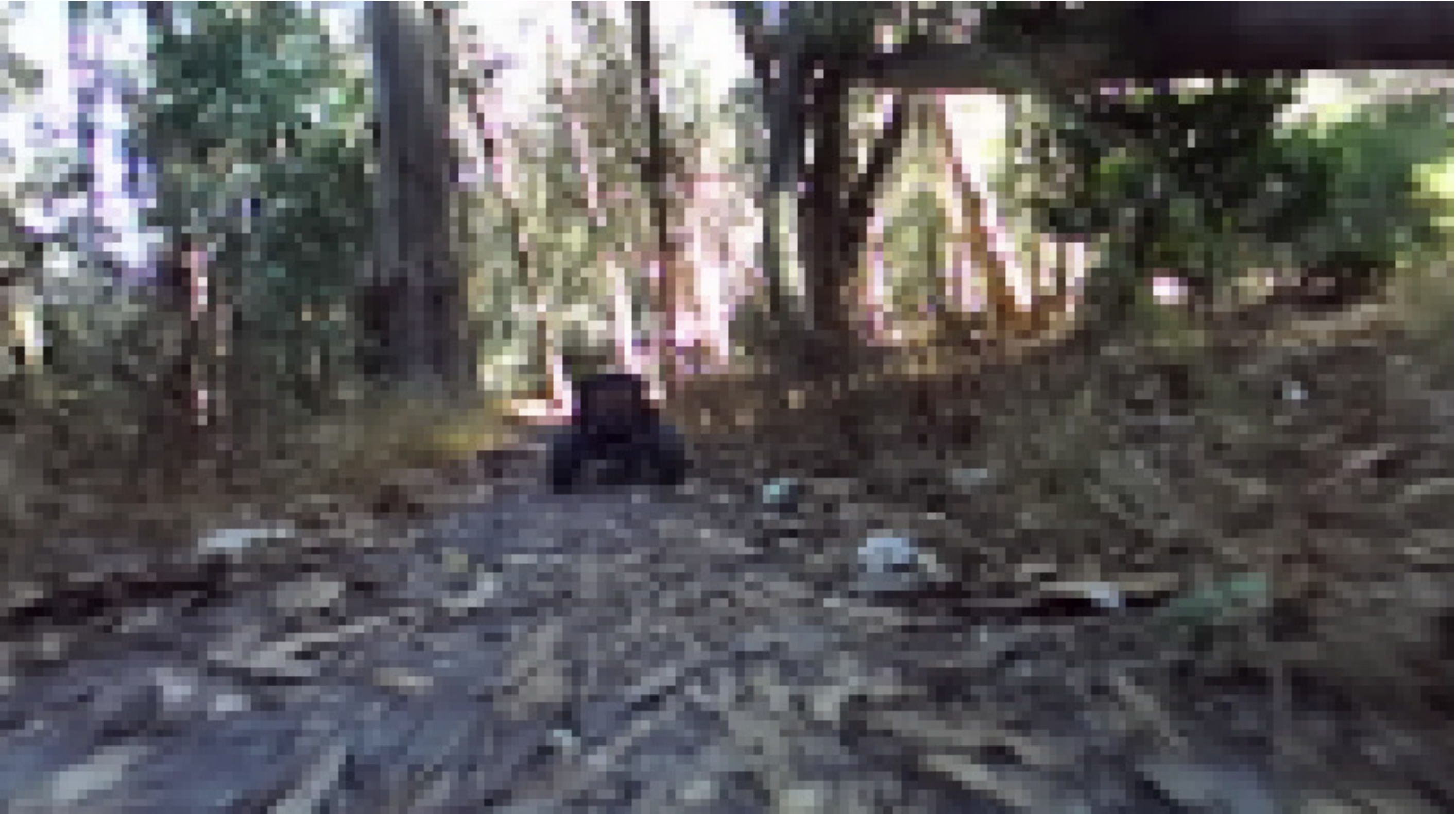}
       \caption{Follow Mode}
       \label{fig1:follow}
    \end{subfigure}
    \begin{subfigure}{0.3\textwidth}
       \centering
       \includegraphics[width=\linewidth]{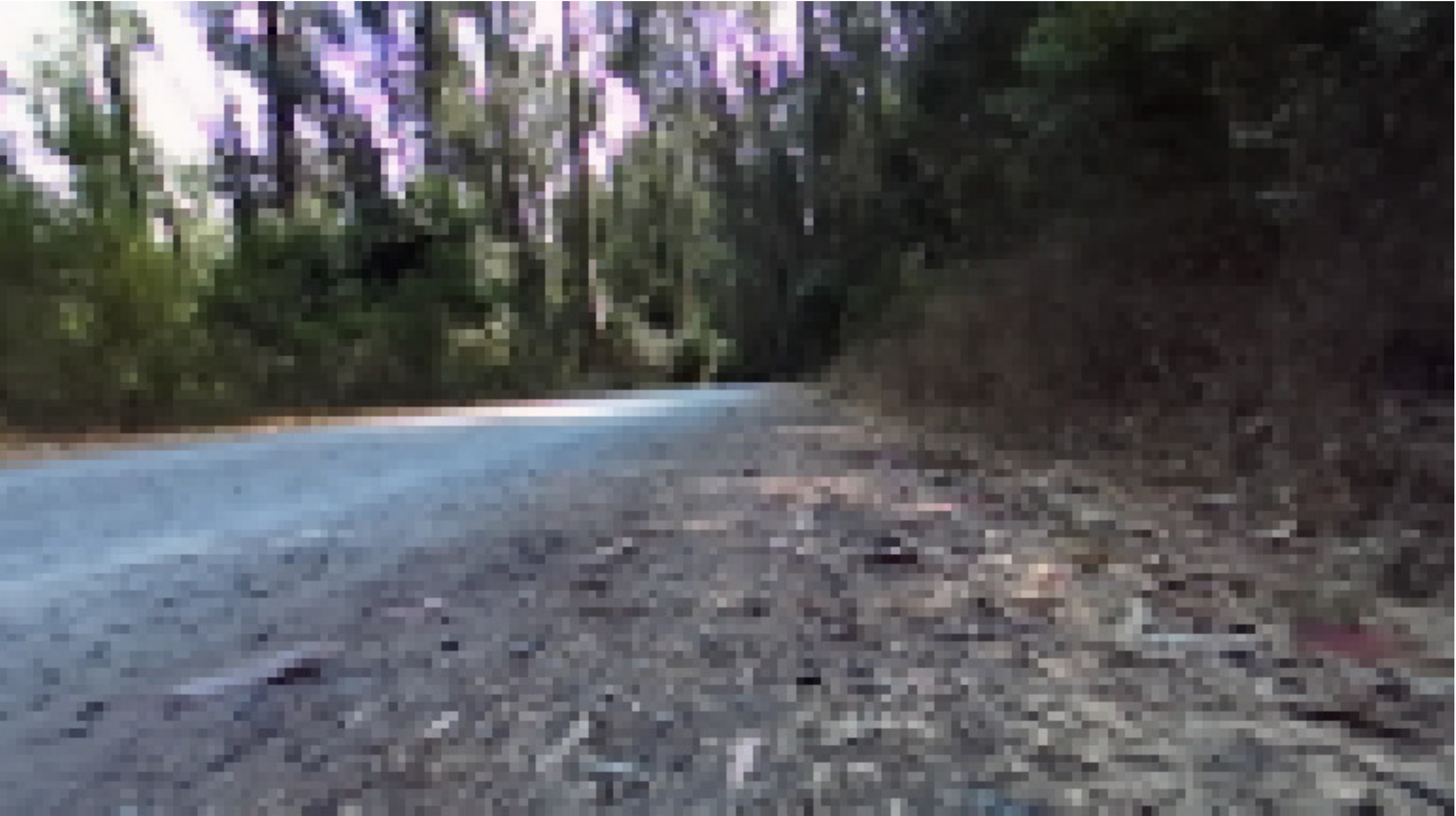}
       \caption{Furtive Mode}
       \label{fig1:furtive}
    \end{subfigure}
    \caption{Behavioral Mode Sample Data from Car's Point of View}
    \label{fig1:behavioralmodes}
\end{figure*}
}

\section{DATASET}
\label{sec:dataset}

While there are many standard datasets available for on-road driving \cite{geiger2013vision,cordts2016cityscapes,huang2018apolloscape,neuhold2017mapillary}, there is an absence of such datasets for unstructured conditions. As part of our research on autonomous driving under unstructured conditions, such as sidewalks, unpaved roads, and trails, we collected driving data using a fleet of 1/10th scale model cars (\Cref{fig:fleet}) similar to that of \cite{intel_paper, williams2017information, giusti2016machine, muller2006off, muller2013real, krasheninnikov2017autonomous}.

Our dataset contains over one hundred hours of driving under diverse geographical, lighting, and weather conditions (\Cref{fig:diverse}). The dataset includes continuous streams of recorded data from a variety of sensors, including stereo images, accelerometer readings, GPS data, steering positions, and motor speed values. Only the raw stereo images, steering, and motor speed values are used in the present work. \Cref{tab:comp} compares our dataset to existing on-road driving datasets.

\subsection{Behavioral Modes}

The dataset also contains annotated modes of behavior, or behavioral modes. We use three distinct behavioral modes:

\begin{enumerate}
    \item \textbf{Direct Mode} consists of data with the car driving with few obstructions or obstacles, usually on a winding sidewalk or forest path (\Cref{fig1:direct}).
    \item \textbf{Follow Mode} consists of data with the car following a lead car in front of it. In this mode, speed modulation occurs as maintaining an uniform distance from the lead car is attempted during driving (\Cref{fig1:follow}).
    \item \textbf{Furtive Mode} consists of data where the car attempts to drive slowly in close proximity to perceived boundaries e.g.\ shrubbery or bushes on either side of a path. If no such boundaries are identified, the car speeds up along the path until one is found (\Cref{fig1:furtive}).
\end{enumerate}

\subsection{Operation Modes}
During data collection runs, we operated the car in one of three operational modes:
\begin{enumerate}
	\item \textbf{Expert Mode} is when the expert driver is in control of the car for the entire data collection process. 
	\item \textbf{Autonomous Mode} is used when evaluating trained networks by allowing a network to infer the speed and steering of the model car.
	\item \textbf{Correctional Mode} is a transient mode. During the autonomous mode, the expert may override the steering or speed controls inferred by the network to recover from mistakes. When this override is engaged, the vehicle momentarily goes into the correctional mode until the expert releases control and autonomous mode is resumed.
\end{enumerate}

Expert and correctional mode data is used for training networks, while autonomous mode is used for evaluating network performance.

\subsection{Dataset Aggregation}
\label{sec:dag}
Our system utilizes imitation learning. Imitation learning has a basic problem, known as "covariate shift", which occurs when a trained network encounters new situations which aren't represented in the dataset of expert driving. In these situations, error compounds quadratically over time to bring the network farther away from expert trajectory \cite{ross2010efficient}.

To solve this problem we implemented a new enhanced approach to the DAgger algorithm \cite{ross2011reduction} which traditionally requires manual labeling of expert trajectories after data are collected from the running network. Instead we make use of recovery data from the correctional operational mode. These new data are then merged with the active dataset for future training in the next iteration. Due to the live corrections, we are able to streamline the data collection process by solving the covariate shift problem while eliminating the need for expert labeling after the data are collected. Our dataset consisted of 19.24\% correctional data and 80.76\% expert data at the time of training and evaluation of the models presented in this paper.

\subsection{Data Moments}

The data collection system gathers time stamps for recorded motor, steer, and camera data. After these data are collected, they are processed, interpolated, and synchronized into packets we call \textit{data moments}. We define a data moment as a set of four RGB input images and an associated collection of ten drive speed and steering angle values. Our networks are trained and evaluated on   series of data moments. A data moment associates the input camera images to motor power and steering angles which when actuated create a spatial trajectory for the car to follow (\Cref{fig:moment}). 

\begin{figure}
\centering
\includegraphics[width=\columnwidth]{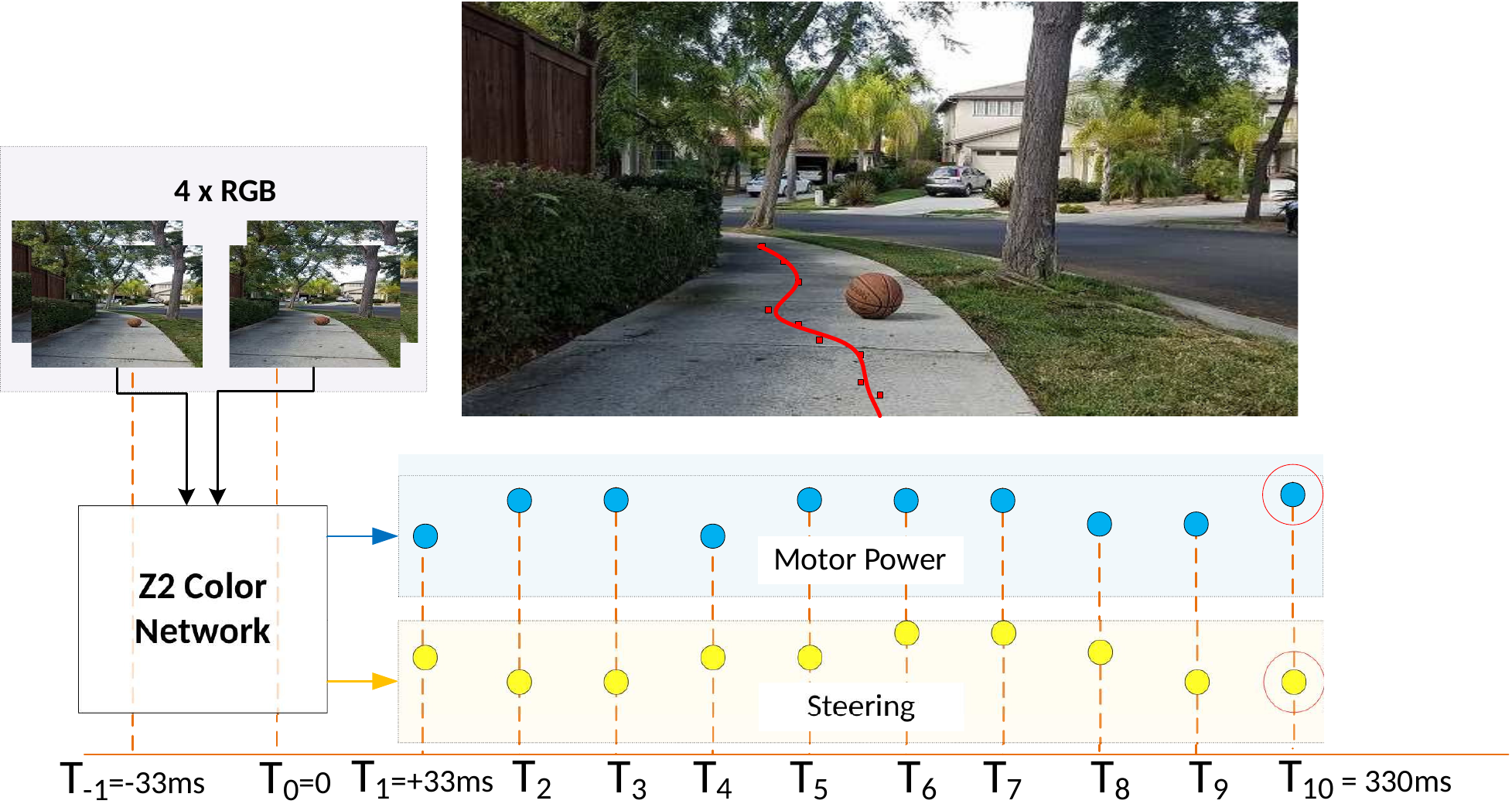}
\caption{Data Moment}
\label{fig:moment}
\end{figure}

For perception of depth, we use left and right images from the stereo camera. For perception of motion, we use image pairs from two time steps -- one image pair is from the current time step, and the other is from 33 ms in the past. This way each data moment contains four RGB images.

The steering and motor values are floating point values ranging from zero to one. In the case of the motor values, one represents a full speed of approximately 9 meters per second. A value of zero for the motors is full speed in reverse; backward driving is used occasionally for abrupt stopping. For steering, a value of one represents the maximum steering angle towards the right, while a value of zero is the maximum steering angle towards the left.

Motor, steering, and image data are collected and stored from the car every 33 ms.
The latency between the network's prediction and actuation on the vehicle is 330 ms to mimic human reaction time. Thus the network predicts 330 ms into the future to account for this delay.

Rather than only training the network to predict a single steering and drive speed value 330 ms into the future, we instead utilize multi-task learning to improve the network's performance through the introduction of the trajectory prediction side task. To accomplish this, we train the networks to predict 10 future time steps, each 33 ms apart. In this case, only the 10th value is used for actuation and inference, while the other values simply serve to improve the car's understanding of the scene during training.

While it is well known that the addition of such side-tasks benefits learning \cite{caruana1998multitask}, we qualitatively confirmed these improvements through on-the-road experiments. In these experiments, networks predicting only final actuation values were compared with MTL networks. It was observed that the MTL networks required far less manual correction and had greater autonomy, suggesting that the side tasks provide the network with improved spatial awareness and driving capability.

\section{METHODOLOGY}
\label{sec:approach}

\begin{figure}[t]
	\centering
	\includegraphics[width=\columnwidth,trim=0 0 20 0,clip]{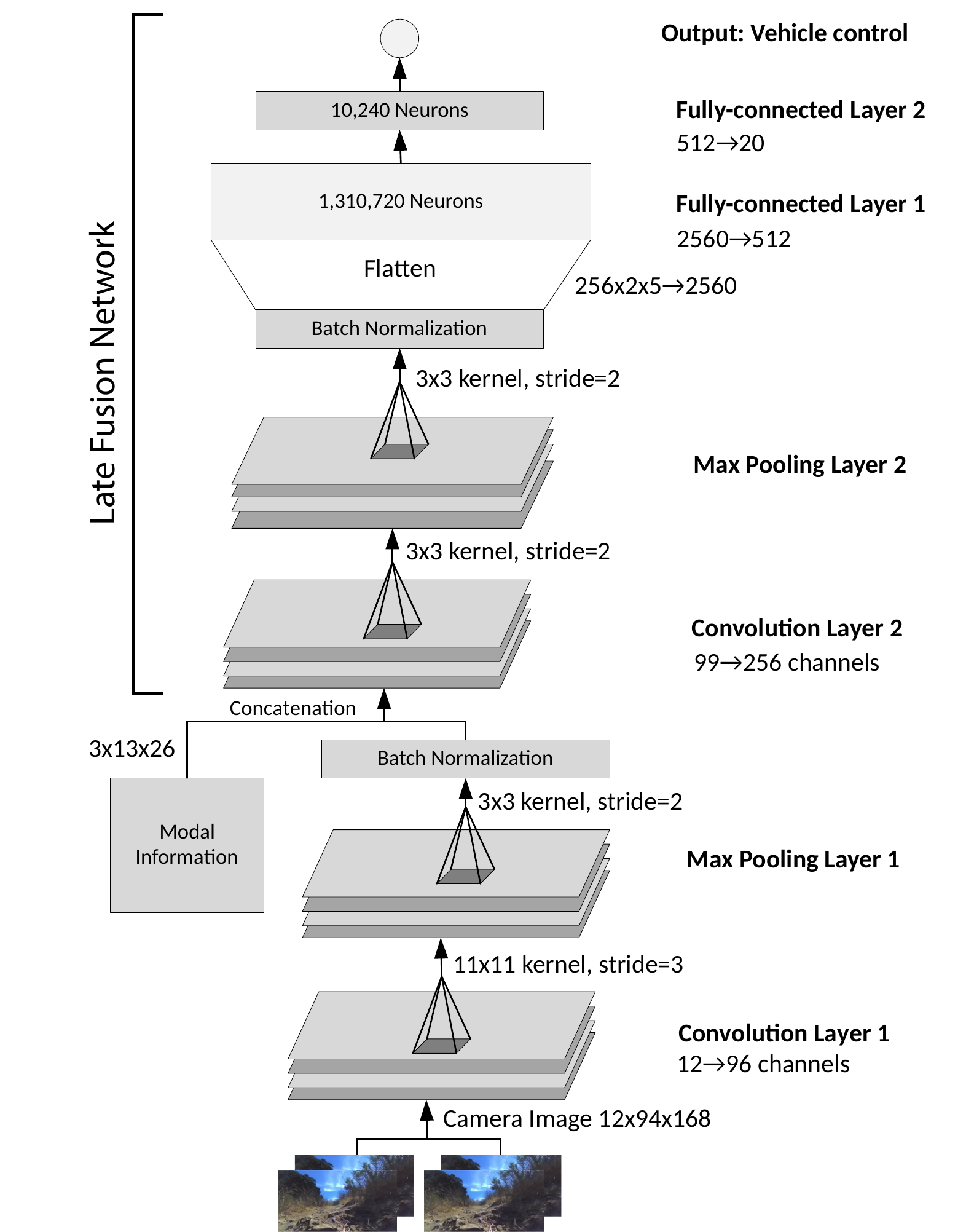}
	\caption{MultiNet Z2Color Network Architecture with Modal Insertion}
	\label{fig:squeeze}
\end{figure}
\subsection{Network Architecture}
For inference, we employ an NVIDIA Jetson TX1 system, and run a custom network we call Z2Color at a 20 Hz frequency. The network consists of two convolutional layers, followed by two fully connected layers shown in (\Cref{fig:squeeze}).

Max pooling and batch normalization were done after each convolutional layer. Max pooling allowed us to efficiently reduce dimensionality and batch normalization prevented internal covariate shift \cite{ioffe2015batch}. The stride and kernel sizes were found empirically through numerous cycles of training and on-road evaluation.
The convolutional layers were designed to act as feature extraction layers, whereas the final fully connected layers act as a steering controller. However, the network was trained in an end-to-end manner so we did not isolate different forms of processing to specific sections of the network.

\subsection{Modal Information}

When collecting data from the cars, along with motor, steering, and image data, we also store the behavioral mode in which the car is being operated. We have trained networks with and without the insertion of the behavioral information and when added, networks more distinctly exhibit individual modal behaviors.

A network without this modal information could potentially learn multiple behavioral modes distinctly, but it would take a great amount of careful training for the filters to separate for each behavioral modality. By adding the modal information within the processing stream, it becomes easier for the network to create independent filters for each behavioral mode.

The behavioral information is inserted as a three channel binary tensor, where each channel represents a different behavioral modality. In order to concatenate with the image going through the convolutional network, the behavioral information is replicated in the spatial dimensions to form a binary tensor of size 3x13x26. The behavioral mode information insertion point in the network was chosen to be after the first convolutional layer in Z2Color (\Cref{fig:squeeze}), allowing for the earlier convolutional layer to generalize basic image processing of the input data without considering behaviors of individual modalities. This replicates the processing of visual data in the macaque monkey in which the early visual cortex receives contextual information from the feedback connections of the frontal cortex from a higher visual cortex. \cite{zipser1996contextual}. Contextual information is not necessary for initial processing of an input image and can thus be inserted later into a deep neural network, as demonstrated by recent works \cite{DBLP:journals/corr/KaiserGSVPJU17, intel_paper}. Ruder  explains that such initial hard parameter sharing reduces the chance of over-fitting by allowing a model to ‘find a representation that captures all of the tasks.’\cite{DBLP:journals/corr/Ruder17a}

\addtolength{\textheight}{-10pt}

\section{EXPERIMENTS}
\label{sec:experiments}

\subsection{Training}
To train our networks, we used the PyTorch\footnote{\url{https://github.com/pytorch/pytorch}} deep learning framework. The networks were trained using the Adadelta Optimizer \cite{DBLP:journals/corr/abs-1212-5701}.

The loss function used for training and validation was Mean Squared Error  (MSE) Loss. During the training phase, the loss was calculated across all values outputted by the network, i.e., across all ten time steps following the formula
\begin{equation} \label{eq:1}
MSE_{train} = \dfrac{1}{2n}(\sum_{t=1}^{n}(s'_t -  s_t)^2 +(m'_t - m_t)^2)
\end{equation}
where $n=10$ is the number of time steps in our case, $s_t$  and $m_t$ are the steering and motor values respectively outputted by the network at a given time step, and $s'_t$ and $m'_t$ are the expert steering and motor values at a given timestep.

During validation, a similar MSE loss metric was used except the loss was calculated only for the two final motor and steering output as given by
\begin{equation} \label{eq:2}
MSE_{validation} = \dfrac{1}{2}((s'_n -  s_n)^2 +(m'_n - m_n)^2)
\end{equation}
Only the final timestep was used in measuring the validation accuracy of the networks as this is the only value which is used for evaluation on the model vehicles, and thus the only value which affects the driving performance of the model car. We chose to use the MSE loss function, as small deviations from expert driving were considered normal while larger deviations are reflective of a problem in the network's control and thus have a greater effect on the calculated error. The quadratic error curve of MSE loss allows for such results and closely mimics results from the percentage autonomy metric introduced in \Cref{sec:careval} in which small deviations have an inconsequential effect on performance.

Each network was trained with the same amount of data. The networks were all evaluated on the same unseen validation set. All experiments were replicated eight times with randomly initialized networks and shuffled datasets. The results here depict the mean across these trials, with error bars representing 95\% confidence intervals.

Our equalized training dataset contains approximately 1.93 million usable data moments for training and validation. 10\% of the collected data were kept for use in an unseen validation dataset for the evaluation of the networks. All data were equally distributed for each modality in both the training and validation sets. 

\begin{figure}[t]
\centering
\includegraphics[width=\linewidth]{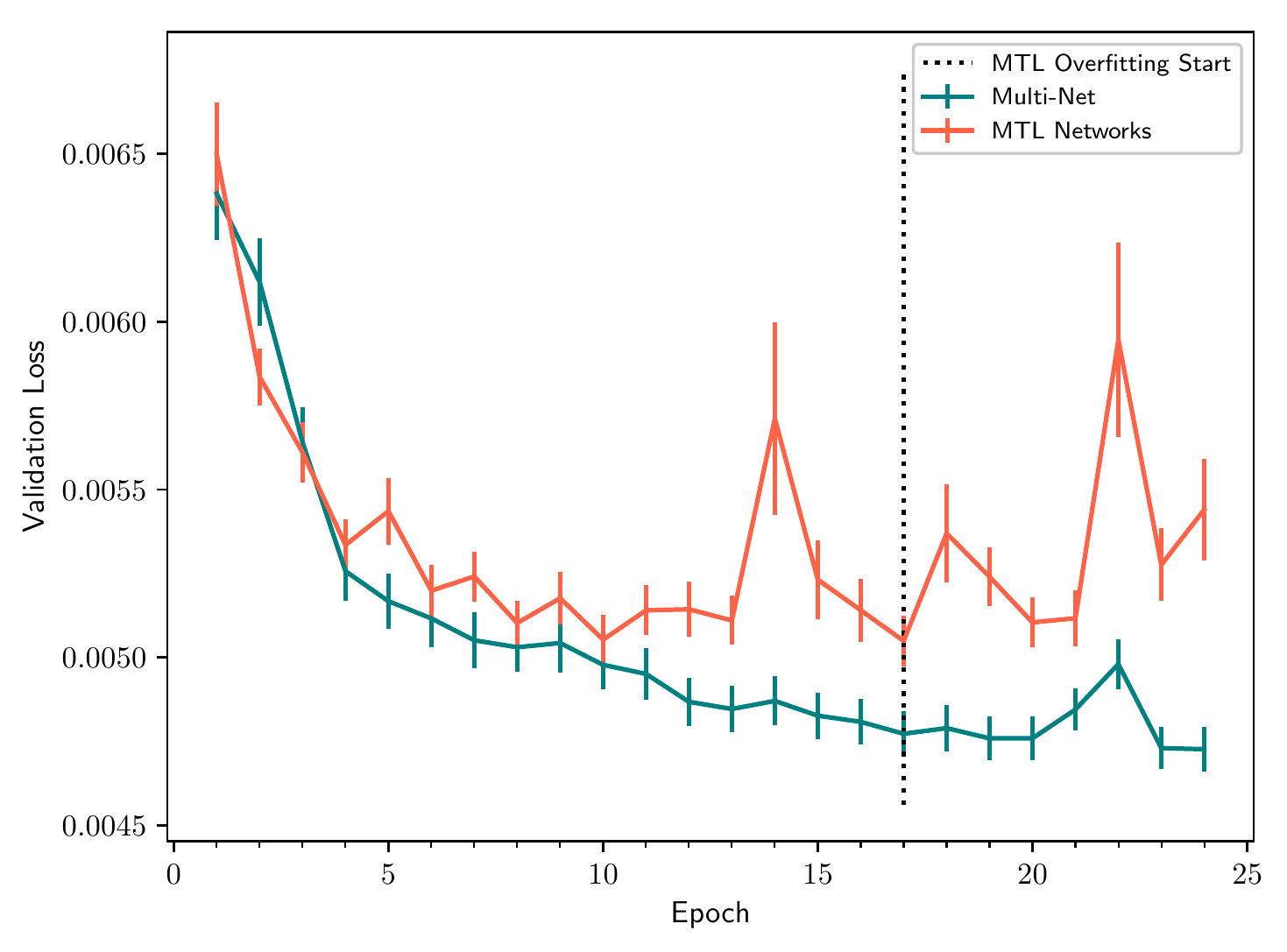}
\caption{Multi-Modal Validation of MultiNet and MTL Networks with 95\% Confidence Intervals}
\label{fig:lve}
\end{figure}

\subsection{Multi-Modal Comparison}
\label{resultspt1}
In our initial experiment a MultiNet Z2Color network trained in a multi-modal dataset of direct, follow, and furtive was compared to three MTL Z2Color networks trained on direct, follow, and furtive modes separately. Each of these networks were trained with an equal number of data moments. The networks were than evaluated using the validation loss measurement described in \Cref{eq:2}. The results are summarized in \Cref{fig:lve} where the losses of the three MTL networks are averaged across the modes for direct comparison to the MultiNet models.

Initially, from epochs 1 to 4, the MultiNets have similar but slightly poorer performance compared to the MTL networks. This is due to the wide variety of data the MultiNets receive requiring greater generalization initially, while the MTL networks can immediately specialize to specific modes.

From epochs 4 to 10, the MultiNets begin to surpass the MTL networks while remaining close in performance. During this period we hypothesize that the MTL networks begin to differentiate between individual driving modalities by using the provided modal information data.

From epochs 10 to 17, the MultiNets drastically outperform the MTL networks, which flatten off in their loss curve here. The MTL loss curve begins to move erratically by getting caught in various local minima. However it doesn't yet begin overfitting, which we characterize as consistently having a loss value above the absolute minimum. The MultiNets steadily improve through the use of the additional modal data. From epochs 17 to 24, the MTL networks begin to overfit dramatically, while the MultiNets continue to decline in loss despite a small bump at epochs 21 and 22. This suggests MTL networks are more susceptible to overfitting and local minima than their MultiNet counterpart. This is likely due to the variety of data the MultiNet networks are exposed to allowing for greater generalization across modes, while maintaining individual behavioral characteristics in specific modalities.

\begin{figure}[t]
\centering
\includegraphics[width=\linewidth]{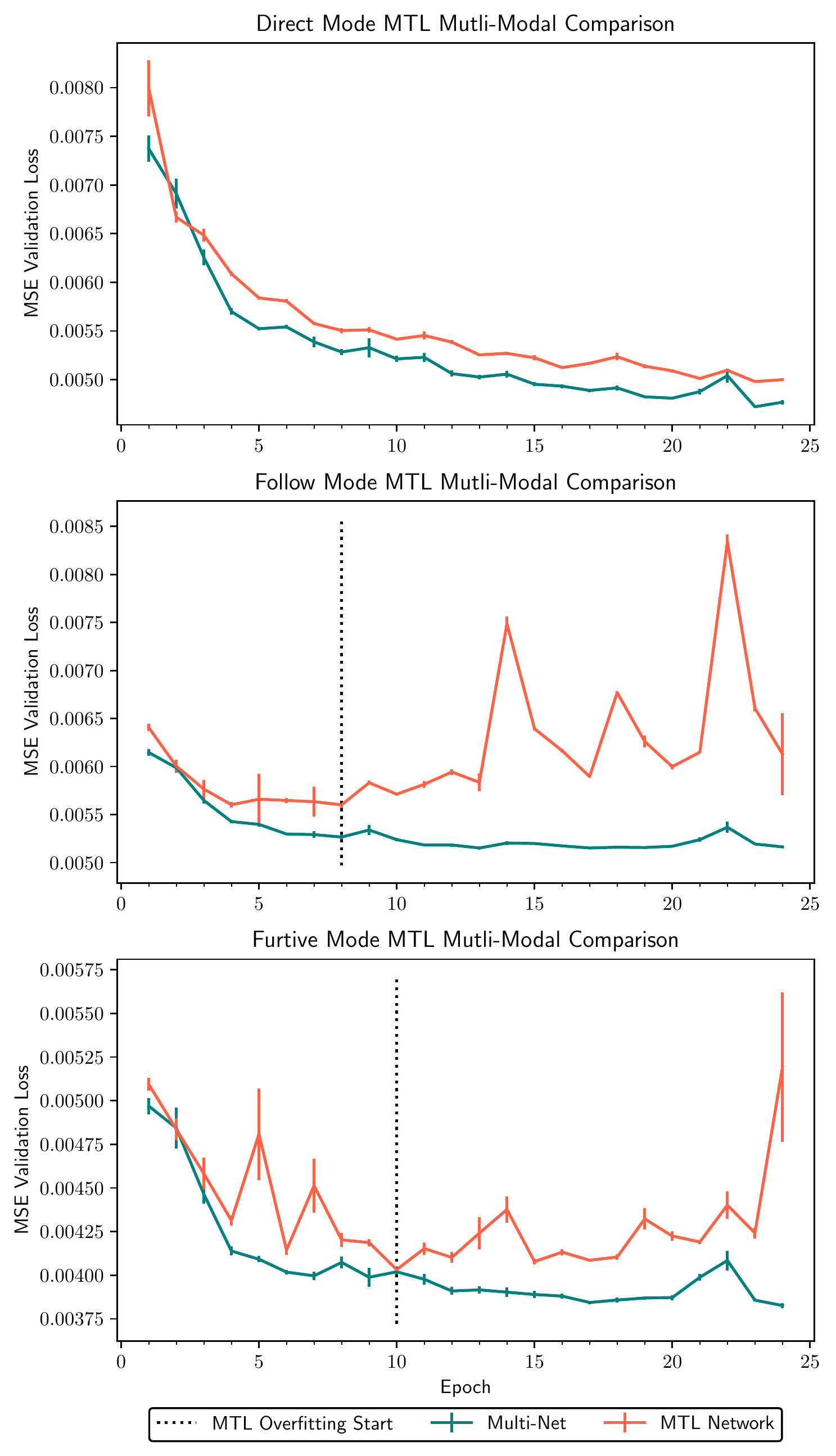}
\caption{Individual Validation of MultiNet and MTL Networks with 95\% Confidence Intervals in Direct, Follow, and Furtive Modes}

\label{fig:furtivegraph}
\end{figure}

\subsection{Performance in Individual Modes}
\label{resultspt2}
To further investigate the network's performance in individual behavioral modes, further experiments were done to compare the MultiNet models to a single MTL network in individual modes. These experiments were conducted to evaluate how the MultiNet displays behaviors distinct to the individual driving modes. In the experiments depicted in \Cref{fig:furtivegraph}, the MultiNet models were trained on Direct, Folow, and Furtive data in each experiment while the MTL networks were trained only on the data for that experiment, e.g., for the furtive graph the MTL network was trained and validated on furtive data. Each network was trained on an equal number of data moments. Validation was done only on the data for the appropriate mode of each network, i.e.\, both the MTL and MultiNet models were evaluated only in furtive mode for the furtive graph. 

In the first mode, direct mode, it is clear that both networks have similar performance levels. In the second epoch, the MTL networks outperformed the MultiNet networks, this is likely due to initial confusion between the behavioral modalities in the MultiNet networks. After this point we hypothesize the MultiNet models learnt the modal distinctions and thus continued to have lower loss measurements than their MTL counterparts until the end of training. Both networks seemed to learn the direct mode task quickly and had no large fluctuations in loss or overfitting as compared to the graphs for each of the other modes. This suggests direct mode is the simplest task as it doesn't involve any special behavioral activity and simply involves avoiding obstacles in the vehicle's path. 

In follow mode, the MTL networks consistently had a higher validation loss than the MultiNet networks. In the second epoch both networks had a similar validation error, but after this point the networks quickly diverged. The MTL networks seemed to flatten out and stop learning significantly after the fourth epoch. By the eighth epoch the follow mode network begins to overfit and steadily increase in loss. Meanwhile, the MultiNet network continues to become more accurate throughout the training other than in two small bumps in the loss in the ninth and twenty-second epoch. These results suggest that MultiNet networks are more capable of learning complex tasks like following, while plain MTL networks are susceptible to overfitting and less capable of learning specific behaviors with the same number of parameters. The results also demonstrate the difference in complexity of the follow mode when compared to the simple direct modality.

In furtive mode, the initial performance in the first four epochs was the same as what was observed for direct and follow mode. It is clear that the multi-modal models seemed to require approximately two epochs of training to begin to distinguish between different modes and develop distinct behaviors. Starting from the fifth epoch, the MTL networks began to oscillate rapidly in the recorded validation loss. This suggests that the MTL networks had found their minima already and were thus oscillating in this region. After the tenth epoch, the MTL models reached their minimum average validation loss. The MultiNet models continued to learn throughout the training period, achieving the lowest validation error at the end of the twenty-four epochs. This suggests the propensity of the MultiNet networks for continuous learning without overfitting as well as for performing in complex behavioral modes like furtive mode.


\subsection{Evaluation on Model Cars}
\label{sec:careval}
To test the proficiency of the models in real world driving situations, we measure the percentage autonomy metric \cite{bojarski2016end} measured as

\begin{equation}
    autonomy = (1 - \dfrac{correction\ time}{elapsed\ time}) \cdot 100
   \label{eq:autonomy}
\end{equation}

\begin{figure}[t]
\centering
\includegraphics[width=\linewidth,,trim=0 0 0 0,clip]{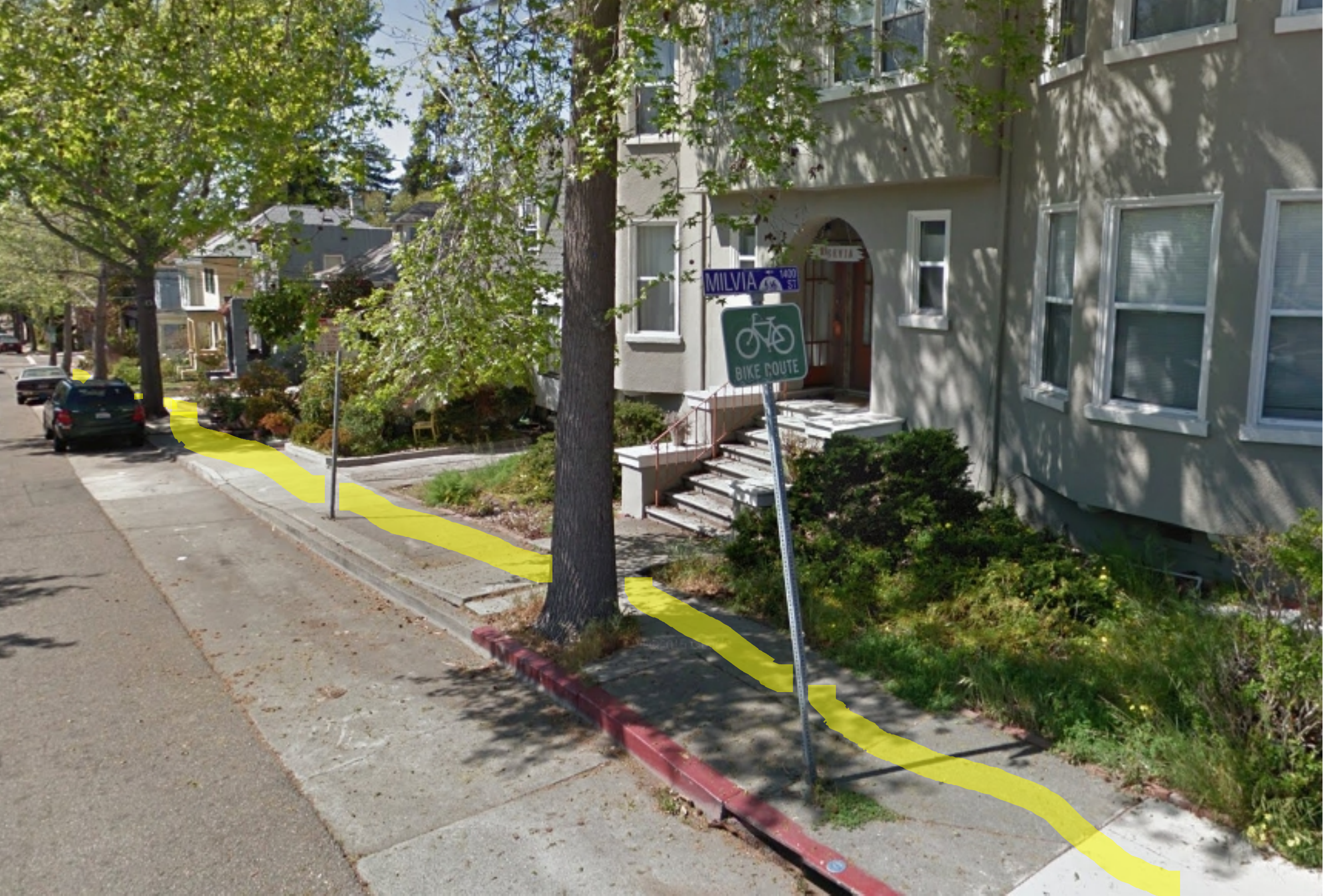}
\caption{Segment of Evaluation Circuit}
\label{fig:evalpath}
\end{figure}

For the live experiments, both MTL and MultiNet networks were evaluated on a winding 200 m loop of sidewalk (\Cref{fig:evalpath}) with sufficient obstacles within a one hour interval. Only direct and furtive modes were used when evaluating the cars, while follow mode was excluded because t he driving of the leader car may differ between runs making a quantitative analysis impractical. The networks for on the road evaluation were chosen at the point of minimum average validation error across the trials, i.e., we chose the epoch and trial which minimized the average validation error for both the MultiNet and MTL networks. This minimum occured at epoch 23 on a specific trial, before either network began to overfit. 

\begin{table}[]
	\begin{center}
		\begin{tabular}{l|l|l|}
			\cline{2-3}
			& \multicolumn{2}{l|}{\textbf{EVALUATION MODE}} \\ \hline
			\multicolumn{1}{|l|}{\textbf{NETWORK}}                   & \textit{Direct}       & \textit{Furtive}      \\ \hline
			\multicolumn{1}{|l|}{\textit{MultiNet}}                  & 92.68\%               & 88.23\%               \\ \hline
			\multicolumn{1}{|l|}{\textit{MTL}}                       & 84.27\%               & 87.55\%               \\ \hline
			\multicolumn{1}{|l|}{\textbf{$\Delta$ (MultiNet - MTL)}} & 8.31\%                & 0.68\%                \\ \hline\hline
			\multicolumn{1}{|l|}{\textbf{$\Delta$ Validation Loss \%}} & 8.16\%                & 12.58\%                \\ \hline
		\end{tabular}
	\end{center}
	\caption{Percentage Autonomy and $\Delta$ Validation Loss \% in Direct and Furtive Modes}
	\label{resultsauto}
\end{table}
The results from our live experiments are depicted in \Cref{resultsauto}. It is apparent that the MultiNet networks were superior to the MTL networks in both evaluation modes. However the difference is more pronounced in the direct mode than the furtive mode. To understand why this is occurring, we have to take a careful look at the autonomy metric which is used to measure performance.

The standard autonomy metric is a good indicator of performance in direct mode, as performance in this mode is solely based on the network's path following and obstacle avoidance abilities. The autonomy metric directly measures this ability.
The furtive mode on the other hand, involves more complex driving behaviors such as speed modulation near foliage, staying close to observed boundaries, in addition to path following and avoiding obstacles. The autonomy metric does not measure these subtle behaviors.
When driving near the edge the chance of going off course is greater and hence the vehicle requires more manual correction. This effectively reduces the autonomy measurement. For these reasons, the standard autonomy metric isn't sufficient when measuring performance in a furtive driving mode in which more complex behaviors than obstacle avoidance are involved.

When observing the MTL and MultiNet networks qualitatively in furtive mode, it is clear that the MultiNet network exhibits more pronounced furtive speed modulation behavior while staying close to boundaries when compared to the MTL network. This characteristic furtive behavior demonstrates the ability of MultiNet networks to act distinctly in multiple behavioral modes with the use of inserted modal data. We have included a supplementary AVI format video which contains segments of footage from each of the live experiments as well as examples of each driving mode. This will be available at \url{http://ieeexplore.ieee.org}.

\subsection{Results Verification}
To verify the results from the validation metric in \Cref{resultspt1} and \Cref{resultspt2}, we computed the percentage difference in performance according to the validation loss to compare the MSE loss metric to the autonomy metric in the live experiments. The percentage difference in validation loss was computed with the following equation:

\begin{equation}
\Delta\ Loss\ \% = \dfrac{MTL\ Loss - MultiNet\ Loss}{MultiNet\ Loss} \times 100
\end{equation}

The $\Delta \ Loss$ \% values for MultiNet and MTL networks in both Direct and Furtive mode are displayed at the bottom of \Cref{resultsauto}. For direct mode, the $\Delta \ Loss$ \% is very similar to the percentage difference in autonomy. This suggests the MSE loss metric is a valid indicator for a network's driving performance. For the furtive mode, the $\Delta \ Loss$ \% is significantly greater than the difference in autonomy which suggests the MSE loss metric used for validation effectively accounts for a network's display of characteristic behaviors in a behavioral mode which were observed qualitatively. This supports the conclusions made earlier from analysis of MultiNet and MTL performance using the MSE loss metric.

\section{CONCLUSION \& FUTURE WORKS}
\label{sec:conclusion}
This paper proposed MultiNet, a methodology for training DNNs to function in several distinct behavioral modes.
This approach inserted behavioral information directly into the network's processing stream, allowing for parameter sharing between related modes.
Additionally, we presented our unique dataset which features over one hundred hours of sidewalk and off-road data recorded from a fleet of model cars driving in diverse conditions.
We tested our MultiNet approach on a reserved validation dataset as well as on the model cars in live experiments.
Our MultiNet approach was shown to exceed the performance of individual networks trained for specific behavioral modes while using fewer parameters.
In our current study, only raw stereo camera images were used as input; future work could integrate additional sensory inputs into the system such as GPS and accelerometer data, which are already available in our dataset.  
Finally, the general MultiNet approach described in this paper can be explored in other applications such as object detection with a modal tensor containing contextual information on the target class.


%


{\small
	\bibliographystyle{ieee}
	\bibliography{egbib}

\begin{thebibliography}{10}\itemsep=-1pt

\bibitem{argyriou2007multi}
A.~Argyriou, T.~Evgeniou, and M.~Pontil.
\newblock Multi-task feature learning.
\newblock In {\em Advances in neural information processing systems}, pages
  41--48, 2007.

\bibitem{bascelli2018sidewalk}
C.~M. Bascelli.
\newblock Where the sidewalk ends and robots deliver: Setting a framework for
  regulating personal delivery devices.
\newblock {\em Rutgers Computer \& Tech. LJ}, 44:93, 2018.

\bibitem{bojarski2016end}
M.~Bojarski, D.~Del~Testa, D.~Dworakowski, B.~Firner, B.~Flepp, P.~Goyal, L.~D.
  Jackel, M.~Monfort, U.~Muller, J.~Zhang, et~al.
\newblock End to end learning for self-driving cars.
\newblock {\em arXiv preprint arXiv:1604.07316}, 2016.

\bibitem{caruana1998multitask}
R.~Caruana.
\newblock Multitask learning.
\newblock In {\em Learning to learn}, pages 95--133. Springer, 1998.

\bibitem{chen2015deepdriving}
C.~Chen, A.~Seff, A.~Kornhauser, and J.~Xiao.
\newblock Deepdriving: Learning affordance for direct perception in autonomous
  driving.
\newblock In {\em Proceedings of the IEEE International Conference on Computer
  Vision}, pages 2722--2730, 2015.

\bibitem{intel_paper}
F.~Codevilla, M.~M{\"u}ller, A.~Dosovitskiy, A.~L{\'o}pez, and V.~Koltun.
\newblock End-to-end driving via conditional imitation learning.
\newblock {\em arXiv preprint arXiv:1710.02410}, 2017.

\bibitem{cordts2016cityscapes}
M.~Cordts, M.~Omran, S.~Ramos, T.~Rehfeld, M.~Enzweiler, R.~Benenson,
  U.~Franke, S.~Roth, and B.~Schiele.
\newblock The cityscapes dataset for semantic urban scene understanding.
\newblock In {\em Proceedings of the IEEE conference on computer vision and
  pattern recognition}, pages 3213--3223, 2016.

\bibitem{eitel2015multimodal}
A.~Eitel, J.~T. Springenberg, L.~Spinello, M.~Riedmiller, and W.~Burgard.
\newblock Multimodal deep learning for robust rgb-d object recognition.
\newblock In {\em Intelligent Robots and Systems (IROS), 2015 IEEE/RSJ
  International Conference on}, pages 681--687. IEEE, 2015.

\bibitem{evgeniou2004regularized}
T.~Evgeniou and M.~Pontil.
\newblock Regularized multi--task learning.
\newblock In {\em Proceedings of the tenth ACM SIGKDD international conference
  on Knowledge discovery and data mining}, pages 109--117. ACM, 2004.

\bibitem{geiger2013vision}
A.~Geiger, P.~Lenz, C.~Stiller, and R.~Urtasun.
\newblock Vision meets robotics: The kitti dataset.
\newblock {\em The International Journal of Robotics Research},
  32(11):1231--1237, 2013.

\bibitem{giusti2016machine}
A.~Giusti, J.~Guzzi, D.~C. Cire{\c{s}}an, F.-L. He, J.~P. Rodr{\'\i}guez,
  F.~Fontana, M.~Faessler, C.~Forster, J.~Schmidhuber, G.~Di~Caro, et~al.
\newblock A machine learning approach to visual perception of forest trails for
  mobile robots.
\newblock {\em IEEE Robotics and Automation Letters}, 1(2):661--667, 2016.

\bibitem{huang2018apolloscape}
X.~Huang, X.~Cheng, Q.~Geng, B.~Cao, D.~Zhou, P.~Wang, Y.~Lin, and R.~Yang.
\newblock The apolloscape dataset for autonomous driving.
\newblock {\em arXiv preprint arXiv:1803.06184}, 2018.

\bibitem{DBLP:journals/corr/HuvalWTKSPARMCM15}
B.~Huval, T.~Wang, S.~Tandon, J.~Kiske, W.~Song, J.~Pazhayampallil,
  M.~Andriluka, P.~Rajpurkar, T.~Migimatsu, R.~Cheng-Yue, et~al.
\newblock An empirical evaluation of deep learning on highway driving.
\newblock {\em arXiv preprint arXiv:1504.01716}, 2015.

\bibitem{ioffe2015batch}
S.~Ioffe and C.~Szegedy.
\newblock Batch normalization: Accelerating deep network training by reducing
  internal covariate shift.
\newblock In {\em International conference on machine learning}, pages
  448--456, 2015.

\bibitem{DBLP:journals/corr/KaiserGSVPJU17}
L.~Kaiser, A.~N. Gomez, N.~Shazeer, A.~Vaswani, N.~Parmar, L.~Jones, and
  J.~Uszkoreit.
\newblock One model to learn them all.
\newblock {\em CoRR}, abs/1706.05137, 2017.

\bibitem{krasheninnikov2017autonomous}
D.~Krasheninnikov et~al.
\newblock Autonomous control of a rc car with a convolutional neural network.
\newblock 2017.

\bibitem{muller2006off}
U.~Muller, J.~Ben, E.~Cosatto, B.~Flepp, and Y.~L. Cun.
\newblock Off-road obstacle avoidance through end-to-end learning.
\newblock In {\em Advances in neural information processing systems}, pages
  739--746, 2006.

\bibitem{muller2013real}
U.~A. Muller, L.~D. Jackel, Y.~LeCun, and B.~Flepp.
\newblock Real-time adaptive off-road vehicle navigation and terrain
  classification.
\newblock In {\em Unmanned Systems Technology XV}, volume 8741, page 87410A.
  International Society for Optics and Photonics, 2013.

\bibitem{neuhold2017mapillary}
G.~Neuhold, T.~Ollmann, S.~R. Bul{\`o}, and P.~Kontschieder.
\newblock The mapillary vistas dataset for semantic understanding of street
  scenes.
\newblock In {\em ICCV}, pages 5000--5009, 2017.

\bibitem{ngiam2011multimodal}
J.~Ngiam, A.~Khosla, M.~Kim, J.~Nam, H.~Lee, and A.~Y. Ng.
\newblock Multimodal deep learning.
\newblock In {\em Proceedings of the 28th international conference on machine
  learning (ICML-11)}, pages 689--696, 2011.

\bibitem{ross2010efficient}
S.~Ross and D.~Bagnell.
\newblock Efficient reductions for imitation learning.
\newblock In {\em Proceedings of the thirteenth international conference on
  artificial intelligence and statistics}, pages 661--668, 2010.

\bibitem{ross2011reduction}
S.~Ross, G.~Gordon, and D.~Bagnell.
\newblock A reduction of imitation learning and structured prediction to
  no-regret online learning.
\newblock In {\em Proceedings of the fourteenth international conference on
  artificial intelligence and statistics}, pages 627--635, 2011.

\bibitem{DBLP:journals/corr/Ruder17a}
S.~Ruder.
\newblock An overview of multi-task learning in deep neural networks.
\newblock {\em CoRR}, abs/1706.05098, 2017.

\bibitem{williams2017information}
G.~Williams, N.~Wagener, B.~Goldfain, P.~Drews, J.~M. Rehg, B.~Boots, and E.~A.
  Theodorou.
\newblock Information theoretic mpc for model-based reinforcement learning.
\newblock In {\em Robotics and Automation (ICRA), 2017 IEEE International
  Conference on}, pages 1714--1721. IEEE, 2017.

\bibitem{DBLP:journals/corr/abs-1801-06734}
Z.~Yang, Y.~Zhang, J.~Yu, J.~Cai, and J.~Luo.
\newblock End-to-end multi-modal multi-task vehicle control for self-driving
  cars with visual perception.
\newblock {\em CoRR}, abs/1801.06734, 2018.

\bibitem{DBLP:journals/corr/abs-1212-5701}
M.~D. Zeiler.
\newblock {ADADELTA:} an adaptive learning rate method.
\newblock {\em CoRR}, abs/1212.5701, 2012.

\bibitem{zhang2012convex}
Y.~Zhang and D.~Yeung.
\newblock A convex formulation for learning task relationships in multi-task
  learning.
\newblock {\em CoRR}, abs/1203.3536, 2012.

\bibitem{zhang2014facial}
Z.~Zhang, P.~Luo, C.~C. Loy, and X.~Tang.
\newblock Facial landmark detection by deep multi-task learning.
\newblock In {\em European Conference on Computer Vision}, pages 94--108.
  Springer, 2014.

\bibitem{zipser1996contextual}
K.~Zipser, V.~A. Lamme, and P.~H. Schiller.
\newblock Contextual modulation in primary visual cortex.
\newblock {\em Journal of Neuroscience}, 16(22):7376--7389, 1996.

\end{thebibliography}
}

\end{document}